# Development of a Neural Network Model for Currency Detection to aid visually impaired people in Nigeria.

Sochukwuma Steven Nwokoye
*School of Science and Technology,*
*Pan-Atlantic University*
Email: snwokoye@pau.edu.ng
(Corresponding author)

Desmond Moru
*School of Science and Technology,*
*Pan-Atlantic University*
Email: dmoru@pau.edu.ng
(Corresponding author)

*Abstract*—Neural networks in assistive technology for visually impaired artificial intelligence's capacity to recognize patterns in complex data. They are used to converting visual data into auditory or tactile representations, helping the visually impaired understand their surroundings. The primary aim of this research is to explore the potential of artificial neural networks to facilitate the differentiation of various forms of cash for individuals with visual impairments. In this study, we built a custom dataset of 3,468 images, which was subsequently used to train an SSD (single-shot detector) neural network model. The proposed system can accurately identify Nigerian cash, thereby streamlining commercial transactions. The performance of the system in terms of accuracy was assessed, and the Mean Average Precision score was over 90%. We believe that our system has the potential to make a substantial contribution to the field of assistive technology while also improving the quality of life of visually challenged persons in Nigeria and beyond.

## I. INTRODUCTION

The combination of innovation and compassion has given rise to transformative solutions that transcend traditional boundaries and redefine the way we address societal challenges in the evolving landscape of technology and artificial intelligence. Among the many challenges that technology has attempted to alleviate, accessibility of financial independence for the visually impaired stands out as a particularly poignant and pressing issue. In Nigeria, a country rich in diversity, heritage, and determination, the visually impaired community faces the formidable challenge of recognizing and managing currency notes. Currency denominations in Nigeria, like many other countries, rely heavily on visual cues, posing an insurmountable challenge to the blind and the visually impaired. Recognizing and managing currency is a critical component of financial independence; yet it remains an elusive aim for this population. This study presents an innovative approach to this difficulty, offering a revolutionary deep learning system tailored solely for cash recognition, with the goal of empowering visually disabled people in Nigeria. This study journey includes the invention, development, and implementation of this modern technology as well as its potential societal and economic implications. The visually impaired community in Nigeria, as in many other countries throughout the world, encounters significant challenges regarding the detection and management of various denominations of money notes. The significance of this challenge cannot be emphasized because it goes beyond ordinary discomfort and strikes at the heart of financial independence and social inclusion. The capacity to distinguish between different money denominations is a key skill in a largely cash-based economy such as Nigeria, allowing citizens to make purchases, pay bills, and engage in various financial transactions. This seemingly simple task is a difficult and frequently insurmountable hurdle for sight impairment. Traditional dependence on visual signals such as size, color, and pattern are ineffective, leaving them open to potential exploitation, fraud, and the inherent discomfort of being unable to manage their funds independently. While efforts have been made to remedy this issue, existing currency identification systems are frequently too expensive for widespread application in developing nations, such as Nigeria, where resources are few. This research was driven by an ardent desire to promote diversity, empowerment, and accessibility. We were motivated by the conviction that technology should serve as a source of hope for those facing obstacles that impede them from reaching their full potential. The goal of this study is to use technology as a change agent, creating a dependable, cost-effective, and scalable solution to enable visually impaired individuals in Nigeria to recognize monetary notes independently. We hope to do this by removing the financial hurdles that have long limited their economic and social participation. We want to provide them with the tools they need to take control of their financial lives, supporting not only economic independence, but also full integration into society as active, self-sufficient members. The purpose of this dissertation is to research the advancement and use of deep learning systems to aid and improve the overall well-being of visually impaired people. This dissertation concentrates on the analysis and implementation of system design within the context of cash detection. In addition, this dissertation investigates the possible effects of deep learning systems on accessibility, inclusivity, and overall user experience. This study also explored the potential of these systems to generate novel possibilities for individuals with disabilities by granting them access to goods and services that were previously inaccessible to them.

*A. Background of Study*

Deep learning systems that combine natural language processing (NLP) and computer vision have shown promise in aiding people with visual impairment. These systems employ deep learning techniques to acquire knowledge representations of data at various levels of abstraction. deep learning techniques have been used in a variety of contexts with the goal of assisting people with visual impairments. Applications include the accurate detection of numerous items as well as the incorporation of sophisticated navigation systems. The deep learning model is trained on a diverse set of photos depicting objects that are highly relevant to those with visual impairments. [1] conducted this study. This paradigm facilitates the effective identification of entities and aids the navigation process.

Visual impairment is a widespread concern that affects many people in various parts of the world. According to data from the World Health Organization, the global population of people with vision impairment is estimated to be 1.1 billion. Notably, approximately 80% of these people live in low- and middle-income countries such as Nigeria (Aniemeka, 2021). According to [2], the estimated prevalence of blindness and visual impairment in Nigerian adults aged 40 years and above is 4.2% and 1.5%, respectively. According to Mamman (2021), the primary causes of avoidable visual impairment in Nigeria include cataracts, glaucoma, refractive errors, harmful traditional eye practices, and corneal opacity.

Visual impairment creates significant barriers to the social and economic well-being of both individuals and communities. Individuals with visual impairment are frequently marginalized and excluded in familial, educational, occupational, and societal contexts (Guardian, 2021). One of the domains in which individuals face challenges is their participation in commercial activities, which is critical for their sustenance and empowerment. Commerce includes a wide range of activities such as commodity and service acquisition, vending, interchange, and commercialization. The ability to discern, quantify, and manage currency as well as discern and distinguish between various products and their corresponding prices is required for the activities.

## II. METHODOLOGY

*A. Data Acquisition*

Because of the nature of this study, there was no readily available dataset to train the model with. Consequently, we had to start from nothing using a custom dataset. A Python script was run to retrieve images from the Bing image search, based on the denomination of currencies as classes for training the model. The quality of the images we obtained using this method was quite variable, so the dataset had to be supplemented by taking our own photographs. Each currency denomination was photographed twice.

A free tool called "labelImg" was used to annotate images. We annotated the images using the Pascal VOC format, which produces an XML file containing information about the objects in the image and their bounding boxes.

However, the model performed well with more data. Therefore, some data augmentation was performed on each of the 346 labeled images to increase the size of the dataset, while also providing a variety to the dataset to avoid overfitting the model. This increased the final total of images in the dataset to 3468 labeled and annotated images.

The images were then divided into training, validation, and test sets. Each set served the following purposes:

- Training: These images were used to train the model. A batch of images from the "train" set was fed into the neural network at each training step. The network predicted the object classes and locations in the images. The training algorithm computes the loss (how "wrong" the predictions are) and uses backpropagation to adjust network weights.
- Validation: The training algorithm can use images from the "validation" set to check the training progress and adjust hyperparameters (such as learning rate). These images, unlike "train" images, are only used intermittently during training (i.e., once every certain number of training steps).
- Test: The neural network never sees these images during the training. They are intended to be used by humans to perform final testing on a model to determine its accuracy.

We created a label map for the detector and converted the images into TFRecords, which is a data format used by TensorFlow for training. Python scripts were run to automatically convert the data into the TFRecord format. A "labelmap.txt" file was created with a list of classes (for example, 10-naira, 50-naira, 100-naira, 200 naira, and so on), adding a new line for each class.

### This creates a "labelmap.txt" file with a list of classes the object detection model
%%bash
cat <<EOF >> /content/labelmap.txt
10 Naira
20 Naira
50 Naira
100 Naira
200 Naira
500 Naira
1000 Naira
EOF

*B. System Architecture*

The proposed solution has a three-tiered system architecture that includes the data acquisition and preprocessing layer, the deep learning model SSD MobileNet V2 FPNLite 320 320 COCO17 TPU-8, and an output processing layer that integrates Google Text-to-Speech.

*1) Data Acquisition and Preprocessing Layer:* This layer involves gathering currency images from various sources and compiling them into a reliable database. The images were then prepared for the model by resizing them to 320× 320 pixels, as required by SSD MobileNetV2 FPNLite, and performing other image preprocessing operations such as normalization, augmentation, and data balance.

*2) Deep Learning Model – Custom trained Tflite SSD MobileNet V2 FPNLite 320x320:* The SSD MobileNet V2 FPNLite 320×320 model was chosen for the task because of its low latency, high-speed performance, and pyramid network, which provides multi-scale object detection, making it ideal for currency detection. The model was pre-trained on the COCO17 dataset and fine-tuned for our specific use case before being converted to Tflite and quantized to run on a device with limited memory and processing power. The model was fed to a processed image with dimensions of 320× 320 pixels. The model employed a feature pyramid network built on top of the MobileNet V2 base architecture. The model was fine-tuned on the preprocessed Nigerian Naira dataset. The currency values were determined using bounding boxes.

*3) Output Processing Layer – Google Text-to-Speech Integration:* The deep learning model's identified currency and denomination of the deep learning model were sent to Google's Text-to-Speech API. This API converts text data into speech, allowing the visually impaired to hear currency denominations. By incorporating the Text-to-Speech API, the solution is transformed into an interactive tool that directly communicates currency denominations to the user.

*4) System Workflow:* • The cash image is transmitted to the preprocessing layer, which performs the appropriate changes.

•The quantized SSD MobileNet V2 FPNLite model is given this modified image, which predicts the class and outputs bounding boxes.

• The deep learning model's identified denomination was then handed on to Google Text-to-Speech for conversion to audio data.

• The auditory output is then relayed to the user, allowing them to determine their money type and value.

This system architecture ensures a clear separation of tasks while fostering strong collaboration among them, allowing for the robust identification and annotation of Nigerian currency denominations in a user-friendly manner for the visually impaired.

*C. Systems Requirements*

The built model was converted into a TensorFlow Lite model, which was designed specifically for running machine learning models on microcontrollers and other devices with limited memory capacity, typically measured in kilobytes. The basic runtime component is designed to consume only 16 KB of memory on an Arm Cortex M3 processor while still allowing it to execute numerous rudimentary models. An operating system, standard C or C++ libraries, or dynamic memory allocation are not required for the implementation.

*D. System Execution/Output*

*1) System Execution for PC:*

*a) Step 1 Download and Install Anaconda:* First, we installed Anaconda, which is a Python environment manager that significantly simplifies Python package management and deployment. Anaconda allows you to create Python virtual environments on a PC without interfering with existing installations of Python. Go to the Anaconda Downloads page and click the download button.

When the download is complete, the download is opened. exe file, and step through the installation wizard. Default installation options were used.

*b) Step 2 Set Up Virtual Environment and Directory:* Go to Start Menu, search for "Anaconda Command Prompt," and click it to open a command terminal. We create a folder called tflite1 directly in C: drive. (You can use any other folder location you like, just to make sure to modify the commands below to use the correct file paths.) Create a folder and move it into it.

Next, create a Python 3.11 virtual environment.

Enter "y" when it asks if you want to proceed. Activate the environment and install the required packages by issuing the following commands: We will install TensorFlow, OpenCV, and a downgraded version of Protobuf. TensorFlow is a large download (approximately 450MB); therefore, it will take a while.

*c) Step 3 Move TFLite Model into Directory:* Next, the custom TFLite model was trained and downloaded from the Colab notebook and moved into the C directory. If you download it from Colab, it should be in a file called custom_model_lite.zip. (If you have not yet trained a model and just want to test it one out, download my "bird, squirrel, raccoon" model by clicking this Dropbox link.) Move the file to C: Directory. Once it has moved, it is unzipped.

At this point, you should have a folder at C: _model_lite, which has at least one detect.tflite and labelmap.txt file.

III. RESULTS

*A. Setting Up and Implementation of Fine-Tuned TensorFlow Object Detection Model*

We have indicated the specific pre-trained TensorFlow model that we intend to utilize from the TensorFlow 2 Object Detection Model Zoo. Additionally, every model is accompanied by a configuration file that indicates file locations, defines training parameters (e.g., learning rate and total number of training steps), and provides other relevant specifications. Finally, we altered the configuration file pertaining to the customized training job. The pre-trained model "ssd-mobilenet-v2-fpnlite-320" was chosen from TensorFlow 2 Model Zoo.

Once the model and configuration files were downloaded, the configuration file was modified by incorporating the high-level training parameters. The variables were used to regulate the progression of the training steps.

• **number_steps :** The total number of steps used to train the model. The appropriate initial value was 40,000 steps.

If the loss metrics continue to decrease after the completion of the training, additional steps should be incorporated. The training duration increased proportionally with the number of steps involved. The training process may be terminated prematurely if the loss function plateaus before reaching the designated number of iterations.

- **batch_size** enumerates the number of images used per training step. A larger batch size facilitates the training of a model with fewer iterations, albeit constrained by the available GPU memory for training purposes. Based on the computational resources provided by the GPUs utilized in the Colab instances, it is advisable to allocate 16 GPUs for SSD models and four GPUs for EfficientDet models.

In this step, additional training information, such as the assigned location of the pre-trained model file, configuration file, and total number of classes, is also determined. Subsequently, the configuration file was modified to incorporate training parameters that were recently specified. The subsequent code segment automatically substitutes the requisite parameters in the downloaded. config file and persists as our customized "pipeline_file.config" file.

The object-detection model recognizes seven unique categories. We used a simple Python script and a Google text- to-speech tool to generate audio feedback files for each class label. Subsequently, these files are saved in the MP3 format with filenames that correspond to each class label. Finally, all the MP3 files were stored in an embedded system subdirectory. The application matches the MP3 file whenever the proposed deep learning model identifies an object. The user was then given audio feedback by playing back the relevant MP3 file. As a result, the proposed system generates audio outputs from the recognized entities.

### B. TensorFlow Lite Model Evaluation and mAP Calculation

*1) Inference test images:* We trained and converted our custom model into the TFLite format. We considered the level of efficacy in detecting objects within images. The images saved in the *test* folder was useful at this point. The absence of test images during the model training process implies that its performance on these images can serve as a reliable indicator of its performance on novel images encountered in real-world scenarios. We developed a code that requires the establishment of a function dedicated to executing inferences on test images. The program initiated the loading process for the images, model, and label map. Subsequently, the model is executed on each individual image, and the obtained results are presented. In addition, the software stores the detection results in the form of text files, enabling us to compute the mean average precision (mAP) score of the model. The following conclusions can be drawn from the data:

inference from Test Images

*2) Calculate mAP:* We needed to quantify our model's accuracy after we had a visual sense of how it was performed on the test photos. "Mean average precision" (mAP) is a prominent approach for quantifying object detection model accuracy. In general, the greater the mAP score, the better the model detects objects in photos. Our test image labeled data and inferenced image test results were relocated to the "ground_truth" and "detection_results" folders, respectively. To calculate the model's accuracy in mAP, the detection results will be compared to the ground truth data.

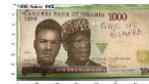

Fig. 1. Inference on 1000 naira

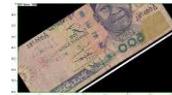

Fig. 2. Inference on 500 naira

*a) Calculate AP for each class:* First, we sorted our TFLite model detection results by decreasing the confidence and assigned them to ground-truth objects. We have "a match" when they share the same label and have an IoU bigger than 0.5 (Intersection over Union greater than 50%). If the ground-truth object has not already been utilized, this "match" is deemed genuine positive.

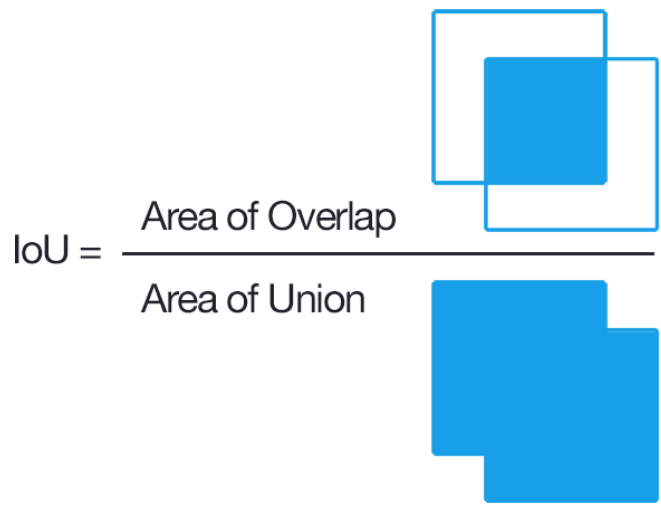

Fig. 3. IoU Formula Diagram

Then, by setting the precision for recall r to the greatest precision achieved for any recall r' > r, we calculated a version of the observed precision/recall curve with monotonically declining precision. Finally, we used numerical integration to obtain the AP as the area under this curve (shown in light blue). Because the curve is piece wise constant, there is no need for approximation.

*b) Calculate mAP:* As indicated in the table above, the model performed well across all the classes. However, because we want a larger perspective, we calculated the mean of all APs, yielding an mAP number ranging from 0 to 100%. As an example:

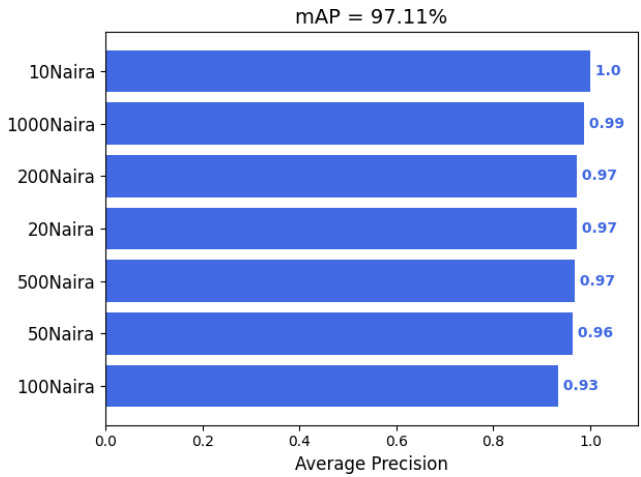

Fig. 4. raph of mAP values for all our classes.

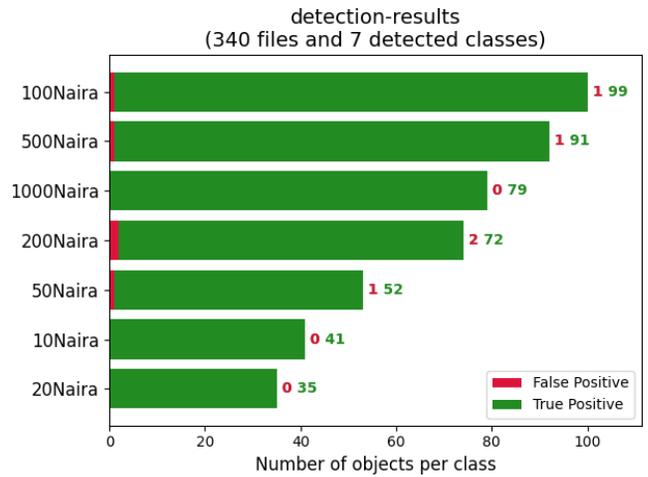

Fig. 5. raph of the True Positive and False Positive of our Predicted Currency.

To cover all the bases, we also calculated the mAP values at other IoU thresholds, which can be discerned from the table below.

TABLE I

| IoU Threshold | Mean Average Precision |
| --- | --- |
| 0.55 | 97.11% |
| 0.60 | 96.97% |
| 0.65 | 96.97% |
| 0.70 | 96.63% |
| 0.75 | 96.50% |
| 0.80 | 96.19% |
| 0.85 | 95.79% |
| 0.90 | 93.29% |
| 0.95 | 83.81% |

*3) True Positives and False Positives:* True Positives (TP) are successfully predicted positive values, indicating that the model accurately forecasted the currency. False Positives (FP) occur when the model predicts currency inaccurately. True Positives and False Positives are essential for comprehending the model's discrimination skills. A histogram plot of our detection_results for TP and FP is shown below.

*4) Log-average Miss Rate (LAMR:* The Log-average Miss Rate(LAMR) is a summary statistic that considers detection performance at various thresholds. It assesses model dependability across a range of false-positive rates. A lower LAMR indicates improved detection performance.

## IV. CHAPTER 4 CONCLUSION

Deep learning systems have emerged as a viable approach for assisting visually impaired individuals using computer vision in an era in which technology is rapidly expanding. These devices use innovative algorithms and artificial intelligence to improve accessibility and independence for people with visual impairments, paving the way for a more inclusive society. These systems can effectively analyze and describe visual information by leveraging the power of deep learning, thereby allowing vision-impaired users to navigate their surroundings with greater ease. Furthermore, advances in computer vision technology have enormous potential for further increasing the functionality and effectiveness of these systems, providing optimism for a future in which visual impairments are no longer a barrier to full participation in society.

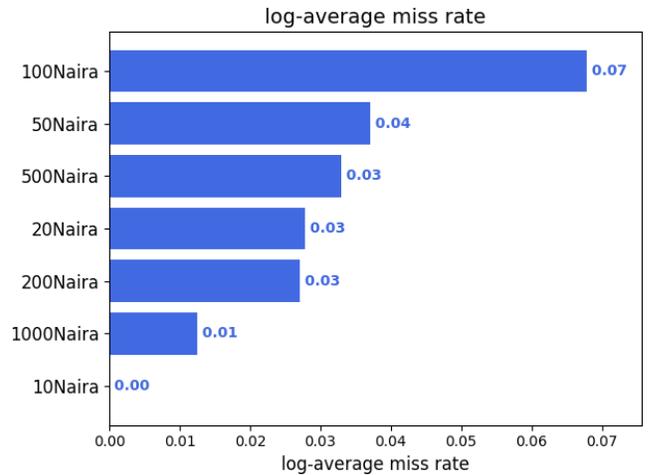

Fig. 6. raph of the Log-average Miss Rate of our Model.

In this thesis, we propose a revolutionary deep learning system that uses natural language processing and computer vision techniques to aid visually challenged persons in Nigeria. We describe the design, implementation, and evaluation of our system, which consists of three main components: a wearable device that captures images and audio from the user's environment; a cloud-based server that processes the data and generates natural language descriptions and instructions; and a speech synthesizer that delivers the output to the user through earphones. We demonstrated that our system could recognize objects, faces, scenes, language, and gestures and deliver helpful information and direction to users in a variety of circumstances. We have also shown that our system can learn from fresh data and adapt to user preferences, needs, and feedback. We compared our system with existing solutions after quantitative and qualitative evaluations. According to the results, our solution outperforms the state-of-the-art methods in terms of accuracy, speed, usability, and user satisfaction. We also reviewed the challenges, limitations, and ethical implications of the system, as well as future work directions. We believe that our system has the potential to make a substantial contribution to the field of assistive technology while also improving the quality of life of visually impaired persons in Nigeria and beyond.

## REFERENCES

[1] R. Joshi, S. Yadav, M. Dutta, and C. González, "Efficient multi-object detection and smart navigation using artificial intelligence for visually impaired people," *Entropy*, vol. 22, no. 9, 2020.

[2] F. Kyari, G. V. S. Murthy, S. Sivsubramaniam, C. Gilbert, M. M. Abdull, G. Entekume, and A. Foster, "Prevalence of blindness and visual impairment in Nigeria: The National Blindness and Visual Impairment Survey," *Investigative Ophthalmology & Visual Science*, vol. 50, no. 5, pp. 2033–2033, 2009.